\documentclass[pmlr]{jmlr}

\usepackage{microtype}
\usepackage{graphicx}
\usepackage{subcaption}
\usepackage{booktabs}
\usepackage{mathtools}

\usepackage[utf8]{inputenc}
\usepackage[T1]{fontenc}
\usepackage{nicefrac}
\usepackage{array}
\usepackage{xcolor}
\usepackage{multirow}
\usepackage{enumitem}

\setlist{nosep}

\newif\ifrevisions
\newcommand{\rev}[1]{\ifrevisions\textbf{\textcolor{red}{#1}}\else#1\fi}

\jmlrproceedings{PMLR}{Proceedings of Machine Learning Research}
\jmlrvolume{340}
\jmlryear{2026}
\jmlrworkshop{Machine Learning for Healthcare}

\title[Explainable Transformers for Clinical Prediction]{Explainable Transformer Models for Clinical Prediction Tasks on Structured Electronic Health Records}

\author{
\Name{Jun Ni Du}\thanks{Equal contribution.}
\Email{jenny.du@sanofi.com}\\
\addr Data \& Computational Science, R\&D, Sanofi\\
Toronto, Canada
\AND
\Name{Lukas Adamek}\footnotemark[1]
\Email{lukas.adamek@sanofi.com}\\
\addr Data \& Computational Science, R\&D, Sanofi\\
Toronto, Canada
\AND
\Name{Maxim Kryukov}
\Email{maksim.kriukov@sanofi.com}\\
\addr Data \& Computational Science, R\&D, Sanofi\\
Barcelona, Spain
\AND
\Name{Flavio Dormont}
\Email{flavio.dormont@sanofi.com}\\
\addr Clinical Real World Evidence, R\&D, Sanofi\\
Cambridge, USA
\AND
\Name{Ziv Bar-Joseph}
\Email{zivbj@cs.cmu.edu}\\
\addr School of Computer Science, Carnegie Mellon University\\
Pittsburgh, USA
\AND
\Name{Sven Jager}
\Email{sven.jager@sanofi.com}\\
\addr Data \& Computational Science, R\&D, Sanofi\\
Frankfurt, Germany
\AND
\Name{Brandon Rufino}\thanks{Corresponding author.}
\Email{brandon.rufino@sanofi.com}\\
\addr Data \& Computational Science, R\&D, Sanofi\\
Toronto, Canada
}

\AtBeginDocument{\let\cite\citep}

\begin{document}

\maketitle

\begin{abstract}
Predictive models over structured electronic health records (EHRs) remain central to machine learning for healthcare, 
but few have jointly emphasized quantitative laboratory information and interpretability with respect to input medical events.
We present BERT-LER, a BERT-style model for coded EHR timelines pretrained and fine-tuned from a de-identified EHR dataset of 75 million patients, that encodes laboratory test results as discrete tokens while retaining graded information through percentile-based binning, paired with Integrated Gradients for token-level attributions grounded in the input EHR sequence.
We benchmark our approach on the public EHRShot benchmark suite and on an asthma severity progression study based on real-world data.
This addresses a methodological gap in EHR foundation-style modeling by unifying laboratory value representation and explainability in a single framework, while assessing whether both predictive performance and explanations generalize beyond standard clinical prediction tasks.
Across EHRShot and asthma tasks, BERT-LER achieves \rev{predictive performance that is competitive with, and on laboratory-related tasks often exceeds, publicly available benchmark models}, and provides attributions that align with clinically known risk factors.
Our architecture and explainability approach can be applied to many therapeutic areas and prediction tasks using language models trained on structured EHRs.
\end{abstract}

\section{Introduction}

Electronic health records (EHR) are central to modern healthcare, capturing longitudinal patient histories including diagnoses, treatments, medications, and laboratory test results. 
These data enable predictive modeling for clinically important tasks such as disease onset, progression, and treatment response, which are critical for decision-making in both clinical care and drug development.
With the move to electronic health records, their use in studies regarding patients and diseases has exploded \cite{Kim2019}. 
Recently, transformer-based models have shown strong performance on EHR prediction tasks by leveraging the sequential and heterogeneous nature of patient records. However, effectively modeling EHR data remains challenging from a machine learning perspective. In particular, structured laboratory measurements are inherently continuous and clinically nuanced, making them difficult to incorporate into discrete token-based architectures without losing important quantitative information. At the same time, clinical deployment requires models whose predictions can be interpreted in terms of patient history, yet many high-performing models provide limited insight into how individual medical events contribute to predictions.
Prior work has explored transformer-based EHR modeling for a range of prediction tasks \cite{ehrshot,transformehr,Rasmy2021medbert,behrt_paper,redekop2025zeroshot,bellamy2024labrador}, 
and some approaches have introduced mechanisms for interpretability, such as attention-based attribution or post-hoc explanation methods \cite{explainable_transformer_ehr,retain}. 
However, prior methods do not typically combine laboratory-aware EHR modeling with token-level interpretability in a way that preserves the graded nature of lab values. 
Moreover, existing interpretability approaches often do not clearly link model predictions back to clinically meaningful input events, limiting their utility for clinical validation and decision support.

In this work, we present BERT-LER (BERT for Lab and Electronic Health Records), a BERT-style model for coded EHR timelines that combines laboratory results tokenization with interpretable predictions. 
We represent laboratory measurements as discrete tokens derived from percentile-based binning, allowing the model to retain graded information while remaining compatible with transformer architectures and token-level explanation methods. To support interpretability, we apply Integrated Gradients to produce token-level attributions that directly link model predictions to input medical events.
We evaluate BERT-LER across two complementary settings. 
First, we benchmark performance on EHRShot \cite{ehrshot}, which is a public benchmark that consists of a longitudinal structured EHR dataset from Stanford Medicine, a released clinical foundation model, and a suite of few-shot prediction tasks.
Second, we conduct an application-driven asthma progression study with explicitly defined inclusion criteria, observation windows, and outcomes, allowing clinicians to assess the model in a realistic setting. This dual evaluation framework supports both methodological comparison and clinical relevance.
Beyond empirical performance, this work provides several generalizable insights for machine learning in healthcare. We show that structured laboratory data can be represented in transformer-based EHR models in a way that preserves clinically meaningful information while remaining compatible with token-level explanation tools. We further show that token-level attribution methods \rev{often reflect} clinically relevant risk factors across both benchmark and application-driven settings. Finally, we highlight how combining strong predictive performance with interpretable outputs may facilitate integration of machine learning models into clinical review and validation workflows.
\rev{The percentile-based laboratory encoding and Integrated Gradients attribution used here build on established techniques; our focus is their integration and evaluation at scale, over a broad structured EHR vocabulary, pretrained on a 75-million-patient corpus and benchmarked across both a public benchmark suite and a prospectively specified clinical protocol.}

\subsection*{Generalizable Insights about Machine Learning in the Context of Healthcare}
The work yields generalizable methodological and analytical takeaways for the ML for health community, not only task-specific scores.
First, structured laboratory measurements can be folded into EHR BERT-style models in a way that preserves graded lab information while remaining compatible with standard token-level explanation tools used in NLP.
Second, combining laboratory-aware modeling with token-level attributions can yield explanations that remain clinically meaningful rather than treating performance and interpretability as separate goals.
Third, evaluating the same approach on both a public benchmark and a prospectively specified asthma cohort study helps connect standard benchmarking with a realistic clinical use case.

\section{Related Work}
\label{section:related_work}

Numerous studies have utilized the encoder-only bidirectional transformer (BERT) model architecture \cite{Rasmy2021medbert,behrt_paper} for clinical applications, including diagnosing previously unidentified diseases, predicting test results, and forecasting treatment outcomes \cite{ehrshot,transformehr,Rasmy2021medbert,behrt_paper}. 
Beyond encoder-only models, generative transformer approaches have also been used to model longitudinal EHR timelines autoregressively, including TransformEHR \cite{transformehr} and a more recent zero-shot GPT study for medical event prediction \cite{redekop2025zeroshot}. 
These studies demonstrate the promise of generative EHR modeling, but the models were trained on substantially smaller EHR cohorts than the de-identified 75-million-patient resource used here (TransformEHR: 6.48 million; zero-shot GPT: 2.4 million).
Another study focused on laboratory data alone, pre-training a transformer on laboratory trajectories and finding limited transfer gains from lab-only modeling \cite{bellamy2024labrador}. 
Taken together, these studies motivate asking whether a laboratory-aware, explainable model pretrained at larger scale can remain competitive against state-of-the-art benchmarks while also supporting a realistic clinical protocol study.
In addition to accurate predictions, clinical researchers emphasize the need for model explainability \cite{SADEGHI2024109370}, as many current deep learning models often fall short in this regard. 
One exception is the work by \citet{explainable_transformer_ehr}, which attributes patient features to specific predictions using a transformer-based model. Similarly, \citet{transformehr} utilized attention scores during model training to identify important features, but this approach does not connect features to patient outcomes. 
Methods like RETAIN \cite{retain} operate on time-series EHR data but provide lower predictive performance compared to attention-based models. More broadly, prior work has not focused on the combination of laboratory-aware modeling and token-level explanations grounded in clinically meaningful input events.

\rev{Several recent studies have explored individual components of laboratory-aware and explainable EHR modeling. For laboratory-value representation, CLMBR \cite{ehrshot}, CHIRon \cite{chiron}, and Labrador \cite{bellamy2024labrador} incorporate laboratory measurements into EHR sequence models, and \citet{binning_joint_embeddings} compare percentile binning against joint continuous--categorical embeddings for numeric EHR values. Labrador is restricted to a small set of laboratory tests within MIMIC and reports limited transfer gains, while the present work models thousands of laboratory codes jointly with diagnoses, medications, procedures, and vaccines. For broader vocabularies and multimodal inputs, Hi-BEHRT \cite{hibehrt} extends BEHRT to long, multimodal sequences but does not provide token-level attribution for laboratory values. For explainability, ExBEHRT \cite{exbehrt} and M-BEHRT \cite{mbehrt} apply gradient-based attribution (expected or integrated gradients) to extended BEHRT models on substantially smaller cohorts ($\sim$10--15k patients), without examining percentile-encoded laboratory values through the same attribution framework. BERT-LER brings laboratory-aware representation and token-level attribution together over a broad structured vocabulary at large scale, evaluated on both a public benchmark and a prospectively specified clinical protocol.}

\section{Methods}
\label{section:methods}

\subsection{Model Pre-training}
\label{section:bert_train}
BERT models were first introduced for natural language, allowing unsupervised pre-training on large unlabelled datasets \cite{Devlin2019BERTPO}. 
This approach was later extended to process structured EHR data and to fine-tune models for disease prediction tasks \cite{Rasmy2021medbert}. 
BERT-LER's pre-training procedure most closely follows the method described in the TransformEHR paper \cite{transformehr}. 
Our model is pre-trained on the pre-training split of the \rev{TriNetX Dataworks} \cite{trinetx} cohort (approximately half of all patients).

Since the data consists of events such as diagnoses, prescriptions, and laboratory tests, the input to the model is a sequence of coded tokens, each representing an event (e.g., an ICD-10 code).
To incorporate lab tests with continuous values, each test result is discretized into one of ten percentile-based categories using the entire \rev{TriNetX Dataworks} dataset as the reference population. 
For example, a value of 1 indicates that the result falls within the 0–10\% percentile range, while a value of 10 corresponds to the 90–100\% percentile range. Please see Appendix~\ref{supp:lab_representation} for how outliers and missing values were handled.

\begin{itemize}
 \item A sequence of input tokens, including demographic information (e.g., age or gender) and medical events in the time window.
 \item A sequence of position IDs, representing the number of days since the first medical event in the time window.
 \item A sequence of percentile IDs, representing the range a given lab test result falls under. Non-lab tokens receive a percentile ID of 0.
\end{itemize}

Each input array is embedded in the model, and the overall embedding for each token is the sum of the token embedding and lab test percentile embedding. Positional embeddings are implemented in the attention mechanism via relative-key-query embeddings, as first described in \citet{shaw-etal-2018-self}.
For each model trained, all medical events in the data are represented by a unique token.
\rev{This factorization of laboratory information into a lab-code embedding and a shared percentile embedding, including its memory advantages over a per-combination vocabulary and its scaling under finer binning, is detailed in Appendix~\ref{supp:lab_representation}.}

Two models were created: our BERT-LER model, which incorporates all diagnosis codes, lab tests, prescriptions, and procedure codes in the data; and a diagnosis-only BERT model, referred to as MED-BERT \cite{Rasmy2021medbert}. 
\rev{For both asthma and EHRShot tasks, we also evaluate a BERT-LER lab-ablation variant that keeps laboratory test tokens but removes percentile-value embeddings, so the model can use lab occurrence timing without access to discretized lab magnitudes.}
A summary of the tokens in each model is shown in Table~\ref{tab:vocab_dist}.
Both models use the BERT transformer architecture introduced in \citet{Devlin2019BERTPO} as implemented in the Hugging Face library \cite{wolf-etal-2020-transformers}. 
The models are pre-trained using masked language modeling (MLM) \cite{Rasmy2021medbert, Devlin2019BERTPO}.
Each model is trained with a hyperparameter search over learning rate and batch size, and a complete overview is provided in Appendix~\ref{supp:model_pretraining}.

\begin{table}[h]
\caption{Number of unique medical codes per ontology for the diagnosis-only BERT (MED-BERT) and BERT-LER models. 
All coding systems are from UMLS \cite{umls}, except demographics, which include age, race, ethnicity, and gender as defined in the \rev{TriNetX} dataset.}
\label{tab:vocab_dist}
\centering
\begin{small}
\begin{tabular}{l|r|r|r}
\toprule
 Ontology & Purpose & BERT-LER & MED-BERT \\
\midrule
ICD10CM & Dx & 65231 & 64497\\
NDC & Rx & 43230 & 0 \\
RXNORM & Rx & 22866 & 0 \\
ICD9CM & Dx & 14136 & 13861 \\
LOINC & Lx & 4680 & 0 \\
CPT & Pr & 1790 & 0 \\
HCPCS & Pr & 1200 & 0 \\
ICD10PCS & Pr & 212 & 0 \\
AGE & Demo & 117 & 113 \\
CVX & Vx & 53 & 0 \\
RACE & Demo & 7 & 7 \\
ETHNICITY & Demo &  3 & 3 \\
GENDER &  Demo & 3 & 3 \\
\bottomrule
\end{tabular}
\end{small}

\end{table}

\subsection{Model Evaluation Methodology}
\label{section:bert_finetune}

We use a mixture of benchmarking tasks to evaluate the performance and explainability of our model. 
Models are fine-tuned for the 14 clinical prediction tasks on EHRShot data \cite{ehrshot}, and population-level explainability is explored. 
In addition, we benchmark our approach on asthma-related clinical prediction tasks to gain insight into disease progression; details of the cohort construction and study design are provided in Appendix~\ref{section:asthma_study_design}. 
In each set of benchmarks, models are fine-tuned for the classification tasks and compared against other approaches, including explainable models such as RETAIN \cite{retain}.

Our asthma disease outcome tasks are defined as:
\begin{itemize}
 \item \textbf{Task 1: Loss-of-control prediction}: For patients classified as having controlled asthma during the baseline period, the model predicts whether they would experience a loss of control (either symptomatic or exacerbation) during the follow-up period. 
 \item \textbf{Task 2: Multiclass prediction}: This task expanded on Task 1 using a three-way classification for remaining controlled, developing symptomatic loss of control, or experiencing an exacerbation. 
 \item \textbf{Task 3: Exacerbation prediction}: For patients symptomatic at baseline, the model predicts whether these patients would experience an exacerbation during the follow-up period.
\end{itemize}
The input to each model consists of data filtered for the relevant task cohort, which is then split into training, validation, and test sets with a 70/15/15 ratio. 
Patients included in the pre-training set (Section~\ref{section:cohort}) are excluded from validation and testing sets for all tasks. 
Table~\ref{tab:split} shows the number of patients in each split and the number of patients in each class. 
To address class imbalance, the model is trained using stratified sampling, ensuring equal class representation in each iteration.
Early stopping with a patience criterion is applied to halt training when performance plateaus or degrades over successive evaluations (ROC-AUC for binary models; average ROC-AUC for multiclass models).
Performance is evaluated using ROC-AUC for all EHRShot benchmark tasks, and both ROC-AUC and PR-AUC for the asthma prediction task. \rev{Confidence-interval estimation by bootstrapping is detailed in Appendix~\ref{supp:bootstrap}.}
For both models, input sequences are limited to 512 tokens or at most five years of patient data; for longer sequences, the earliest events are removed.

% \begin{table}[t]
% \caption{Number of patients in each split for each prediction task. 
% Numbers in parentheses represent the number of patients in the negative/positive class for binary tasks and the controlled/exacerbation/symptomatic class for the multiclass task.}
% \label{tab:split}
% \centering
% \begin{small}
% \begin{tabular}{l|l|l|l}
% \toprule
% Prediction Task & Train & Validation & Test \\
% \midrule
% Loss-of-control & 19575 (11612/7963) & 4143 (2422/1721) & 4108 (2472/1636) \\
% Multiclass & 19575 (11612/1168/6795) & 4143 (2422/270/1451) & 4108 (2472/239/1397)\\
% Exacerbation & 50438 (45496/4942) & 10748 (9727/1021) & 10859 (9774/1085) \\
% \bottomrule
% \end{tabular}
% \end{small}

% \end{table}

%% New formatting to fit in one column
\begin{table}[h]
\caption{Number of patients in each split for each prediction task. 
Parentheses show class counts: negative/positive for binary tasks and controlled/exacerbation/symptomatic for multiclass.}
\label{tab:split}
\centering
\begin{small}
\begin{tabular}{l|l|l}
\toprule
Prediction Task & Split & Patients (class counts) \\
\midrule
Loss-of-control & Train & 19575 (11612/7963) \\
               & Validation & 4143 (2422/1721) \\
               & Test & 4108 (2472/1636) \\
\midrule
Multiclass     & Train & 19575 (11612/1168/6795) \\
               & Validation & 4143 (2422/270/1451) \\
               & Test & 4108 (2472/239/1397) \\
\midrule
Exacerbation  & Train & 50438 (45496/4942) \\
               & Validation & 10748 (9727/1021) \\
               & Test & 10859 (9774/1085) \\
\bottomrule
\end{tabular}
\end{small}

\end{table}

\subsection{Model Interpretability}
\label{section:bert_explain}
\begin{figure}[h]
 \centering
 \includegraphics[width=0.6\linewidth]{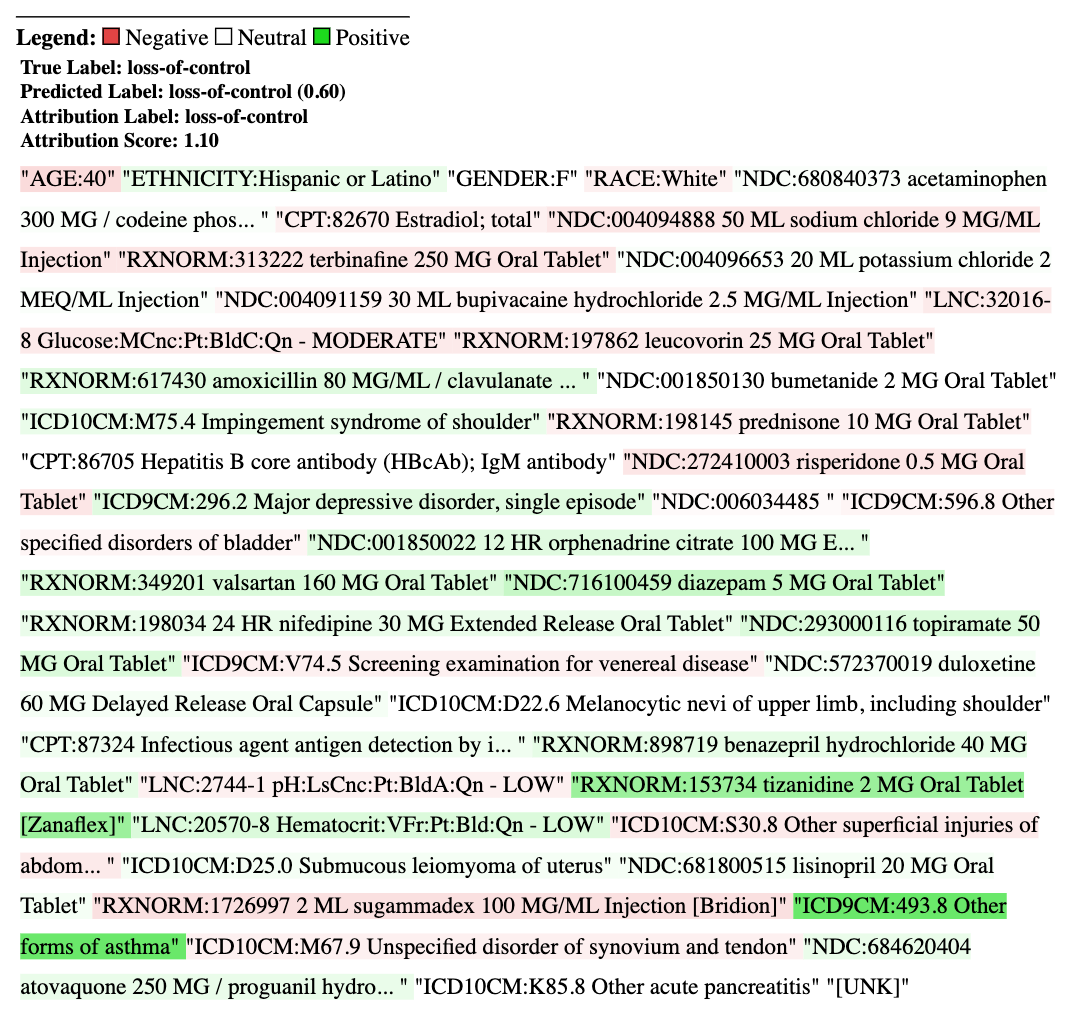}
 \caption{Distribution of attribution scores for a hypothetical patient in the loss-of-control model. 
The model predicted a 60\% probability of loss of control. 
Input tokens are shown on the right, with darker green highlights indicating stronger positive contributions to the prediction.}
 \label{fig:single_patient_viz}

\end{figure}

As mentioned in the Introduction, a critical need for clinical applications is the ability to explain model predictions and tie them to specific input attributes. To achieve this, we utilized the Integrated Gradients (IG) method \cite{axiomatic_attribution}, which provides a systematic approach to quantify the contribution of each input token to a particular prediction. 
\rev{IG is a post-hoc attribution method applied to an already-trained model rather than a component of the architecture; it does not alter the model's parameters or predictions. We use IG in preference to perturbation-based methods such as SHAP or counterfactual (input-deletion) analysis primarily for scalability: those methods evaluate the model over many feature subsets or single-token deletions and scale with input-sequence length, whereas the cost of IG scales with the number of integration steps and is independent of sequence length, which is advantageous for EHR sequences that can span years of history.}
The IG method attributes the difference between the model’s prediction for an actual input and a baseline input \cite{axiomatic_attribution}.
The baseline serves as a neutral reference point to highlight the contribution of features and is typically chosen to represent an "uninformative" input state, such as a vector of all zeros \cite{axiomatic_attribution}.
\rev{Note that for lab tests, attributions are run on the combined embedding of the lab token and the percentile embedding, indicating the combined contribution of the lab test and its value to a prediction.}
Given these two inputs, IG computes the gradient of the model's prediction along a path interpolating between the baseline and the actual input. 
In our case, the baseline and actual input are represented as embeddings: the baseline is an all-zeros vector, and the actual input is represented by its corresponding embedding. The attribution score for each feature is calculated by summing over the hidden dimension.
Specifically, the integrated gradient for the \(i\)-th input feature is defined as: \cite{axiomatic_attribution}
\[
IG_i(x) = (x_i - x'_i) \int_{\alpha=0}^{1} \frac{\partial F(x' + \alpha (x - x'))}{\partial x_i} d\alpha
\]
Here, \(F\) denotes the model’s prediction function; 
\(x' + \alpha (x - x')\) represents the interpolated input with scaling parameter \(\alpha\), 
and \(\frac{\partial F(x)}{\partial x_i}\) is the gradient of \(F(x)\) along the \(i\)-th dimension. 
In practice, the integral over \(\alpha\) is approximated using a Riemann sum over \(m\) steps, 
where \(\alpha = \frac{k}{m} \text{ for } k = 1, \ldots, m\) \cite{axiomatic_attribution}. 
The term \((x_i - x'_i)\) scales this average gradient by the input change to reflect the attribution of the \(i\)-th feature within the embedding space. 
The output vector has the same shape as the token embedding, and a scalar attribution score is obtained by summing across all dimensions. 
For our implementation, we used the open-source \mbox{\textit{transformers-interpret}} package \cite{Pierse_Transformers_Interpret_2021}, 
which relies on PyTorch's explainability package Captum \cite{paszke2017automatic, kokhlikyan2020captum}.
Modifications were made to adapt the method for EHR data, rather than its original application in natural language processing tasks.
Sequential position IDs were replaced with relative dates of events, and token type IDs, originally used to distinguish sentences, were repurposed to represent percentiles of lab values.
Figure~\ref{fig:single_patient_viz} visualizes the attribution score distribution for a hypothetical patient. Attribution scores are calculated for each token for every patient in the test set. 
Additionally, we extend the scoring function to downweight infrequent features, helping to emphasize the more informative features; see Appendix~\ref{section:bert_attribution_aggr} for details.

\subsection{Comparison Methods}

We benchmark our model against prior approaches to EHR-based prediction and simpler count-based models.
For the EHRShot tasks, we benchmarked against the pre-trained CLMBR model \cite{ehrshot}, RETAIN \cite{retain}, and Med-BERT \cite{Rasmy2021medbert}. Simpler models, such as XGBoost and logistic regression, were already benchmarked against CLMBR in \cite{ehrshot}.
\rev{A breakdown of the released CLMBR vocabulary by source ontology is provided in Appendix~\ref{supp:clmbr_vocab} for comparison with the BERT-LER vocabulary in Table~\ref{tab:vocab_dist}.}
For our asthma benchmarking tasks, we compare against RETAIN \cite{retain}, MED-BERT \cite{Rasmy2021medbert}, and a set of simpler count-based models. 
\rev{CLMBR is not included on the asthma tasks because the released model vocabulary codes diagnoses predominantly in SNOMEDCT\_US, whereas our cohorts code diagnoses in ICD10CM and ICD9CM, so applying it would require substantial concept remapping or retraining (Appendix~\ref{supp:clmbr_vocab}). The included asthma baselines (RETAIN, XGBoost, and logistic regression) all have access to the same laboratory value information as BERT-LER, differing only in representation (Appendix~\ref{supp:count_comparison}).}
Details on these models, including their training procedures and hyperparameter tuning, are provided in Appendix~\ref{supp:count_comparison}.

\section{Cohort}
\label{section:cohort}

\subsection{Cohort Selection}

For pre-training and our custom benchmark, we use de-identified EHR data from \rev{TriNetX Dataworks} \cite{trinetx}, 
which provides longitudinal records including diagnoses, procedures, medications, and laboratory values for approximately 75 million patients. A detailed description of data governance, de-identification, and regulatory compliance is provided in Appendix~\ref{section:data_governance}. 
We split the \rev{TriNetX Dataworks} cohort into two roughly equal disjoint sets: one for model pre-training and one for downstream benchmarking, ensuring that no pre-training patients appear in evaluation cohorts. Pre-training uses medical events after 2012 for patients with at least two events separated by 365 days. 
We then constructed the asthma cohorts for our custom benchmarking task using only the remaining set, ensuring that no patients involved in pre-training were used in downstream testing. 

To evaluate generalizability, we benchmark on the 14 clinical prediction tasks from the EHRShot dataset \cite{ehrshot}. 
EHRShot is a publicly available de-identified longitudinal EHR dataset from Stanford Medicine containing 6,739 patients with coded timelines of demographics, diagnoses, procedures, medications, and laboratory results. We follow the benchmark protocol and report results using the canonical train/validation/test splits provided with the release.

\subsection{Data Extraction}

Models consume sequences of coded tokens representing demographics and medical events in chronological order; laboratory values enter through percentile-based discretization as described in Section~\ref{section:bert_train}. For EHRShot, coding systems differ from our pre-training source (e.g., SNOMEDCT\_US in EHRShot versus ICD10CM for diagnoses in pre-training); we use UMLS to map concepts into the fine-tuning vocabulary when evaluating on EHRShot (Section~\ref{subsection:ehrshot_benchmarking}).

\section{Results}

\subsection{Evaluation Approach/Study Design}

We evaluate BERT-LER and baselines under two complementary designs. On EHRShot \cite{ehrshot}, we use the canonical train/validation/test splits and report ROC-AUC for 14 classification tasks, following the public benchmark protocol. On the asthma severity progression tasks, we use cohorts and labels defined in Section~\ref{section:bert_finetune} and Appendix~\ref{section:asthma_study_design}, with a 70/15/15 patient-level split, stratified training, early stopping, and ROC-AUC plus PR-AUC as in Section~\ref{section:bert_finetune}. 
We benchmark our explainable language model on these two application settings: EHRShot as a standard multi-task benchmark, and prospectively specified asthma tasks as an external application check beyond public leaderboards.

\subsection{EHRShot Benchmarking}
\label{subsection:ehrshot_benchmarking}

\begin{table}[t]
\caption{Average ROC-AUC values of various models on the EHRShot prediction tasks. Bold values indicate the best-performing model for each task and metric. Parentheses show the 95\% confidence interval bounds derived from bootstrapping, reported as (upper, lower). \rev{We additionally report a BERT-LER lab-ablation variant with lab tokens retained but percentile-value embeddings removed.}}
\label{ehrshot_rocs}
\centering
{\scriptsize
\setlength{\tabcolsep}{3pt}
\resizebox{\linewidth}{!}{%
\begin{tabular}{c|c|c|c|c|c}
\toprule
BERT-LER & \rev{\shortstack{BERT-LER\\(no lab value emb.)}} & CLMBR & Med-BERT & RETAIN & Prediction Task \\
\midrule
0.80 (0.85, 0.75) & \rev{0.76 (0.81, 0.70)} & \textbf{0.84 (0.88, 0.80)} & 0.69 (0.75, 0.63)& 0.65 (0.72, 0.58) & ICU Admission \\
\textbf{0.80 (0.83, 0.77)} & \rev{0.79 (0.82, 0.76)} & 0.79 (0.82, 0.76) & 0.72 (0.75, 0.69) & 0.70 (0.74, 0.66) & Readmission to ICU \\
0.78 (0.80, 0.76) & \rev{0.77 (0.79, 0.75)} & \textbf{0.80 (0.82, 0.78)} & 0.69 (0.71, 0.67) & 0.71 (0.74, 0.69) & Long Stay in ICU \\
\textbf{0.910 (0.912, 0.907)} & \rev{0.787 (0.790, 0.783)} & 0.891 (0.893, 0.889) & 0.688 (0.692, 0.684) & 0.800 (0.804, 0.796) & Anemia Lab \\
\textbf{0.79 (0.80, 0.78)} & \rev{0.70 (0.71, 0.69)} & 0.77 (0.78, 0.76) & 0.68 (0.69, 0.66) & 0.63 (0.65, 0.61) & Hyperkalemia Lab \\
\textbf{0.77 (0.78, 0.76)} & \rev{0.65 (0.67, 0.63)} & 0.75 (0.76, 0.74) & 0.63 (0.64, 0.61) & 0.58 (0.60, 0.57) & Hypoglycemia Lab \\
\textbf{0.867 (0.870, 0.864)} & \rev{0.698 (0.704, 0.693)} & 0.733 (0.737, 0.72) & 0.63 (0.64, 0.62) & 0.59 (0.60, 0.58) & Hyponatremia Lab  \\
\textbf{0.808 (0.811, 0.805)} & \rev{0.801 (0.804, 0.799)} & 0.783 (0.786, 0.781) & 0.708 (0.712, 0.704) & 0.728 (0.732, 0.724) & Thrombocytopenia Lab \\
0.73 (0.77, 0.69) & \rev{0.75 (0.79, 0.71)} & 0.74 (0.78, 0.70) & \textbf{0.75 (0.79, 0.71)} & 0.59 (0.64, 0.53) & Acute MI Diagnosis \\
0.60 (0.68, 0.51) & \rev{0.57 (0.69, 0.44)} & \textbf{0.70 (0.81, 0.57)} & 0.57 (0.72, 0.41) & 0.66 (0.76, 0.56) & Celiac Disease Diag. \\
0.67 (0.71, 0.63) & \rev{0.68 (0.73, 0.64)} & \textbf{0.69 (0.73, 0.64)} & 0.65 (0.69, 0.61) & 0.53 (0.57, 0.49) & Hyperlipidemia Diag. \\
\textbf{0.73 (0.77, 0.69)} & \rev{0.68 (0.73, 0.64)} & 0.70 (0.74, 0.66) & 0.65 (0.69, 0.60) & 0.55 (0.60, 0.50) & Hypertension Diag. \\
\textbf{0.81 (0.89, 0.72)} & \rev{0.78 (0.90, 0.66)} & 0.69 (0.80, 0.55) & 0.75 (0.83, 0.66) & 0.47 (0.59, 0.36) & Lupus Diag. \\
0.80 (0.87, 0.73) & \rev{0.84 (0.89, 0.77)} & 0.84 (0.90, 0.78) & \textbf{0.85 (0.90, 0.80)} & 0.70 (0.78, 0.62) & Pancreatic Cancer Diag. \\
\bottomrule
\end{tabular}}
}

\end{table}

The EHRShot dataset was created to benchmark EHR models on various clinical tasks \cite{ehrshot}. Since the coding systems in \rev{TriNetX} and EHRShot differ, for example, ICD10CM is used in \rev{TriNetX} for diagnosis and SNOMEDCT\_US in EHRShot, we use UMLS to translate medical concepts in EHRShot to those available in MED-BERT's and BERT-LER's vocabulary \cite{umls}. 
We present the average ROC-AUC for each task in Table~\ref{ehrshot_rocs}. We include benchmarks against RETAIN \cite{retain} and CLMBR \cite{ehrshot} trained on the original, untranslated data. Our BERT-LER model shows superior performance in most tasks.
BERT-LER consistently improves performance on all lab test prediction tasks, taking into account statistical uncertainties. It also performs significantly better for predicting lupus diagnoses. 
For other diagnosis tasks, the performance of CLMBR and BERT-LER is comparable within statistical uncertainties. Performance on outcome-related tasks, such as ICU admission, readmission, and long stays, is also comparable between CLMBR and BERT-LER. 
There are no tasks with MED-BERT or RETAIN better than CLMBR or BERT-LER, when taking into account statistical uncertainties.

\subsection{EHRShot Explainability: Lab Anemia Tests}

\begin{figure}[t]
 \centering
 \includegraphics[width=0.8\linewidth]{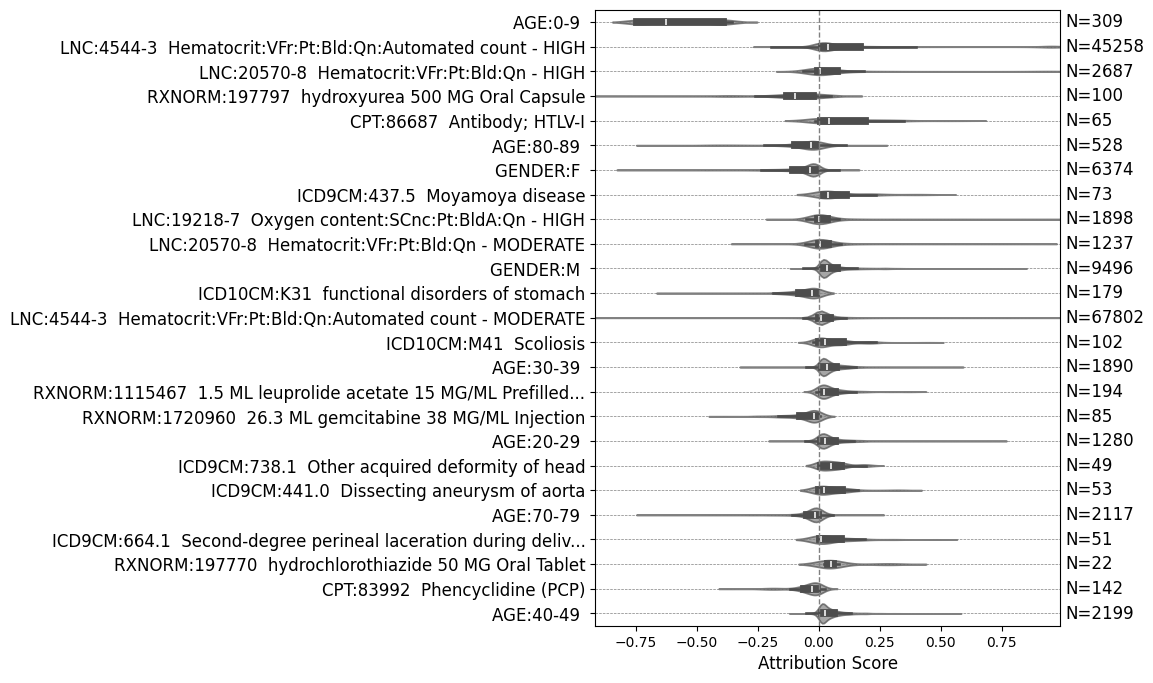}
 \caption{Attribution score distribution for normal lab-anemia test results on the EHRShot dataset. Each row corresponds to a unique token. Black bars show the 68\% interval, the white line indicates the median, and the numbers on the right (N) show the token counts in the test set.}

 \label{fig:lab_anemia_clmbert}

\end{figure}

Figure~\ref{fig:lab_anemia_clmbert} shows our explainability results for predicting normal lab anemia test results. Because attribution is directional, features with negative contributions to the normal class can also be interpreted as contributing toward abnormal anemia-related results. 
In terms of age, children and the very elderly are found to be at higher risk of abnormal anemia-related results compared to other age groups, consistent with other sources \cite{anemia_common_children, anemia_common_elderly}. The model also assigns women, compared to men, more to being at higher risk of abnormal anemia-related results, consistent with existing evidence \cite{anemia_common_women}. 
The main lab features associated with normal test results include high red blood cell concentration tests 
%(LNC:4544-3 and LNC:20570-8) 
and tests related to blood oxygen content.
%(LNC:19218-7)% 
These are consistent with known markers of normal anemia-related lab results \cite{lab_tests_and_anemia}. 
In terms of diagnoses, functional disorders of the gut attributes towards abnormal lab test results, consistent with known connections between gut disorders and acquired anemia \cite{lab_anemia_and_gut}. 
% We did not find evidence linking scoliosis or Moyamoya disease with anemia. 
For prescriptions, hydroxyurea is found to be strongly attributed towards low anemia values; however, there is limited supporting evidence in the literature \cite{hydroxyurea_anemia}. A few chemotherapy agents, such as gemcitabine, are also attributed to lab anemia.
We provide a comparison figure from the RETAIN method \cite{retain} in Figure~\ref{fig:retain_lab_anemia} in the Appendix. 
Compared to our method, RETAIN lacks many of the notable features captured by our model. None of RETAIN’s top features are clearly linked to anemia-related outcomes.

\subsection{Asthma Benchmarking}
To evaluate our approach, we benchmarked our model on three asthma-related prediction tasks: 
loss of control, multiclass disease state, and exacerbation prediction, as described in Section~\ref{section:bert_finetune}. 
Table~\ref{tab:performance_metrics} summarizes the PR-AUC and ROC-AUC results across all evaluated models. 
Across all tasks and both metrics, BERT-LER consistently achieves the strongest performance, indicating robust predictive capability relative to both deep learning and traditional machine learning baselines.
\rev{The lab-ablation BERT-LER variant (lab tokens retained, percentile-value embeddings removed) performs between MED-BERT and full BERT-LER across tasks, indicating that laboratory value magnitude contributes additional predictive signal beyond lab event presence alone.}

For the loss-of-control task, BERT-LER attains the highest PR-AUC and ROC-AUC, outperforming MED-BERT, RETAIN, and count-based approaches. 
Gradient-boosted trees ranked second overall, while logistic regression and RETAIN exhibit comparatively lower discrimination. 
Similar trends are observed in the multiclass prediction task, where BERT-LER again achieves the best overall performance, with XGBoost providing competitive PR-AUC.
The exacerbation task exhibited a distinct performance profile, with lower PR-AUC across all models, reflecting the inherent class imbalance and the difficulty of predicting acute events. 
Despite this, BERT-LER substantially outperforms other approaches, achieving the highest PR-AUC and ROC-AUC. XGBoost ranks second, while MED-BERT, RETAIN, and logistic regression demonstrate more limited performance.
These gains are consistent with BERT-LER’s ability to leverage richer longitudinal information and larger training cohorts, which may be particularly important for modeling rare but clinically significant outcomes.

\begin{table}[h]
\caption{Performance of the models across three prediction tasks for asthma. 
Bold values indicate the best-performing model for each task and metric. 
Values in parentheses represent 95\% confidence intervals derived from bootstrapping.}
\label{tab:performance_metrics}
\centering
\begin{small}
\begin{tabular}{l|c|c}
\toprule
Prediction Task & PR-AUC & ROC-AUC \\
\midrule
\textbf{Loss-of-control} & & \\
BERT-LER & \textbf{0.599 (0.572, 0.624)} & \textbf{0.682 (0.665, 0.698)} \\
\rev{BERT-LER (no lab value emb.)} & \rev{0.586 (0.558, 0.611)} & \rev{0.676 (0.660, 0.692)} \\
Med-BERT & 0.509 (0.484, 0.533) & 0.622 (0.604, 0.640) \\
RETAIN & 0.476 (0.452, 0.501) & 0.593 (0.577, 0.612) \\
XGBoost & 0.554 (0.528, 0.581) & 0.647 (0.630, 0.665) \\
Logistic & 0.482 (0.457, 0.506) & 0.594 (0.578, 0.611) \\
\midrule
\textbf{Multiclass} & &  \\
BERT-LER & \textbf{0.627 (0.611, 0.643)} & \textbf{0.667 (0.653, 0.682)} \\
\rev{BERT-LER (no lab value emb.)} & \rev{0.609 (0.593, 0.625)} & \rev{0.649 (0.636, 0.663)} \\
Med-BERT & 0.562 (0.545, 0.578) & 0.606 (0.592, 0.620) \\
RETAIN & 0.516 (0.501, 0.529) & 0.544 (0.532, 0.556) \\
XGBoost & 0.611 (0.594, 0.629) & 0.656 (0.640, 0.671)\\
Logistic & 0.560 (0.543, 0.576) & 0.603 (0.586, 0.619) \\
\midrule
\textbf{Exacerbation} & &  \\
BERT-LER & \textbf{0.232 (0.211, 0.255)} & \textbf{0.724 (0.708, 0.739)} \\
\rev{BERT-LER (no lab value emb.)} & \rev{0.213 (0.195, 0.235)} & \rev{0.706 (0.691, 0.722)} \\
Med-BERT & 0.183 (0.166, 0.202) & 0.665 (0.649, 0.681) \\
RETAIN & 0.170 (0.154, 0.186) & 0.650 (0.632, 0.667) \\
XGBoost & 0.204 (0.185, 0.224) & 0.688 (0.672, 0.704) \\
Logistic & 0.180 (0.164, 0.198) & 0.666 (0.649, 0.684)\\
\bottomrule
\end{tabular}
\end{small}

\end{table}

\subsection{Asthma Explainability}

\begin{figure}[t]
 \centering
 \includegraphics[width=0.8\linewidth]{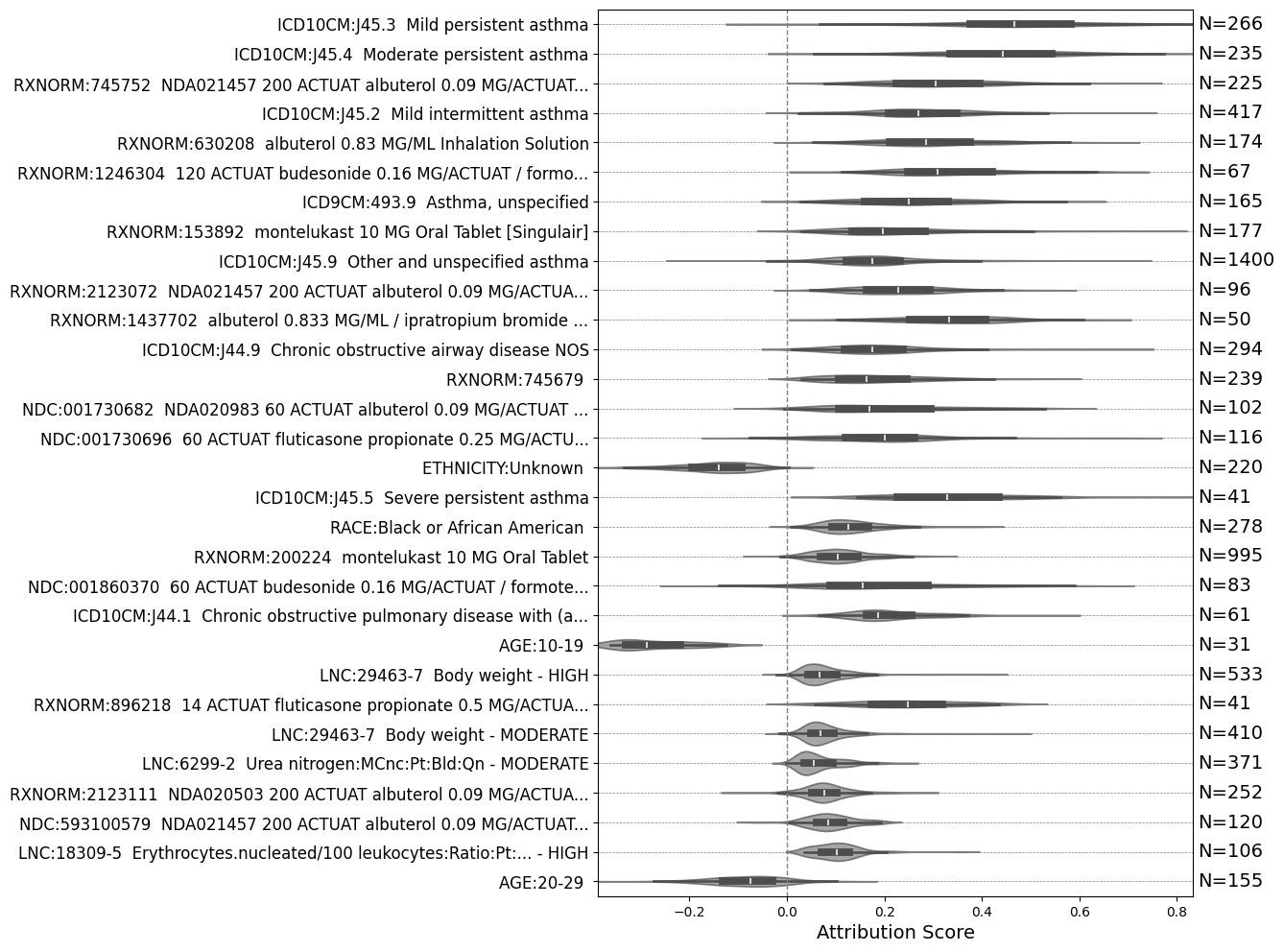}
 \caption{Attribution score distribution for predicting asthma loss of control. Top 30 tokens with minimum frequency $\geq 30$ are shown, ordered by adjusted scores.}
 \label{fig:overall_scores}

\end{figure}

Our model not only predicts the likelihood of worsening asthma but also produces a ranked list of the features that contribute most strongly to those predictions. 
Figure~\ref{fig:overall_scores} shows the top predictors, ordered by their adjusted attribution scores. 
Most of the highest-ranked features align closely with clinical expectations and with top contributing factors identified by RETAIN and the XGBoost model (Figures~\ref{fig:retain_top_features} and~\ref{fig:xgboost_top_features} in the Appendix).
Asthma severity codes are among the strongest predictors, including mild intermittent, mild/moderate/severe persistent, and unspecified asthma.
This pattern suggests that any level of persistent or symptomatic asthma signals insufficiently controlled disease, consistent with findings from comparison models.

Medication-related features are also highly ranked. 
Frequent use of short-acting beta-agonists (SABAs), such as albuterol formulations, emerges as a strong predictor of future asthma worsening. 
Increased SABA use often reflects unstable day-to-day asthma control and has been consistently associated with a higher risk of worsening \cite{saba}. 
\rev{A single underlying medication can appear under several distinct tokens; for example, albuterol inhalers are represented by multiple RxNorm and NDC codes reflecting differences in prescribing and pharmacy coding practices. Several separately ranked medication tokens in Figure~\ref{fig:overall_scores} may therefore correspond to the same clinical agent, and their attributions are best interpreted jointly rather than as independent signals.}
Several long-acting beta-agonists (LABAs) combined with inhaled corticosteroids (ICS), such as budesonide–formoterol and several fluticasone propionate formulations, also appear prominently among the top features. 
Patients requiring LABA--ICS combinations typically have more severe disease or inadequate control on ICS alone, both of which are associated with greater susceptibility to future worsening \cite{laba_ics}. 
Montelukast use also appears among the top predictors, again reflecting ongoing controller therapy needs in patients with atopic or persistently symptomatic asthma. 
Similar tokens are consistently found among the top contributing features for the RETAIN and XGBoost models.

Comorbidity-related features appear as well. Chronic obstructive pulmonary disease (COPD) is highly ranked, consistent with the overlap between asthma and COPD and the increased risk of severe respiratory events \cite{asthma_copd}.
Elevated or abnormal laboratory markers, such as high nucleated erythrocyte counts and altered urea nitrogen/creatinine ratios, also appear as predictors. 
Although these markers do not map directly onto classic asthma comorbidities, they are associated with systemic stress and severe respiratory conditions, including acute respiratory distress \cite{nrbc_respiratory_distress} and the prediction of acute exacerbations in COPD \cite{urea_nitrogen_creatinine_copd}.

Demographic factors also play a meaningful role.
Black or African American race is associated with increased risk of asthma worsening, consistent with extensive prior research documenting disparities in asthma outcomes \cite{african_american}. 
Age-related signals also emerge, with the 10--19 age group showing a \rev{negative} association relative to other age categories, aligning with studies reporting that adolescents generally experience fewer asthma exacerbations than young children or older adults \cite{age_asthma}. 
Conversely, unknown ethnicity is associated with slightly reduced risk, though this may reflect data capture patterns rather than a true biological effect.

Finally, obesity-related features, represented by high or moderate body weight, are also associated with a higher future risk of worsening. 
Obesity exacerbates airway inflammation and increases asthma morbidity, in agreement with prior literature \cite{obesity} and comparison models.

\subsubsection{Asthma Lab Test Explainability}

\begin{figure}[t]
 \centering
 \includegraphics[width=0.8\linewidth]{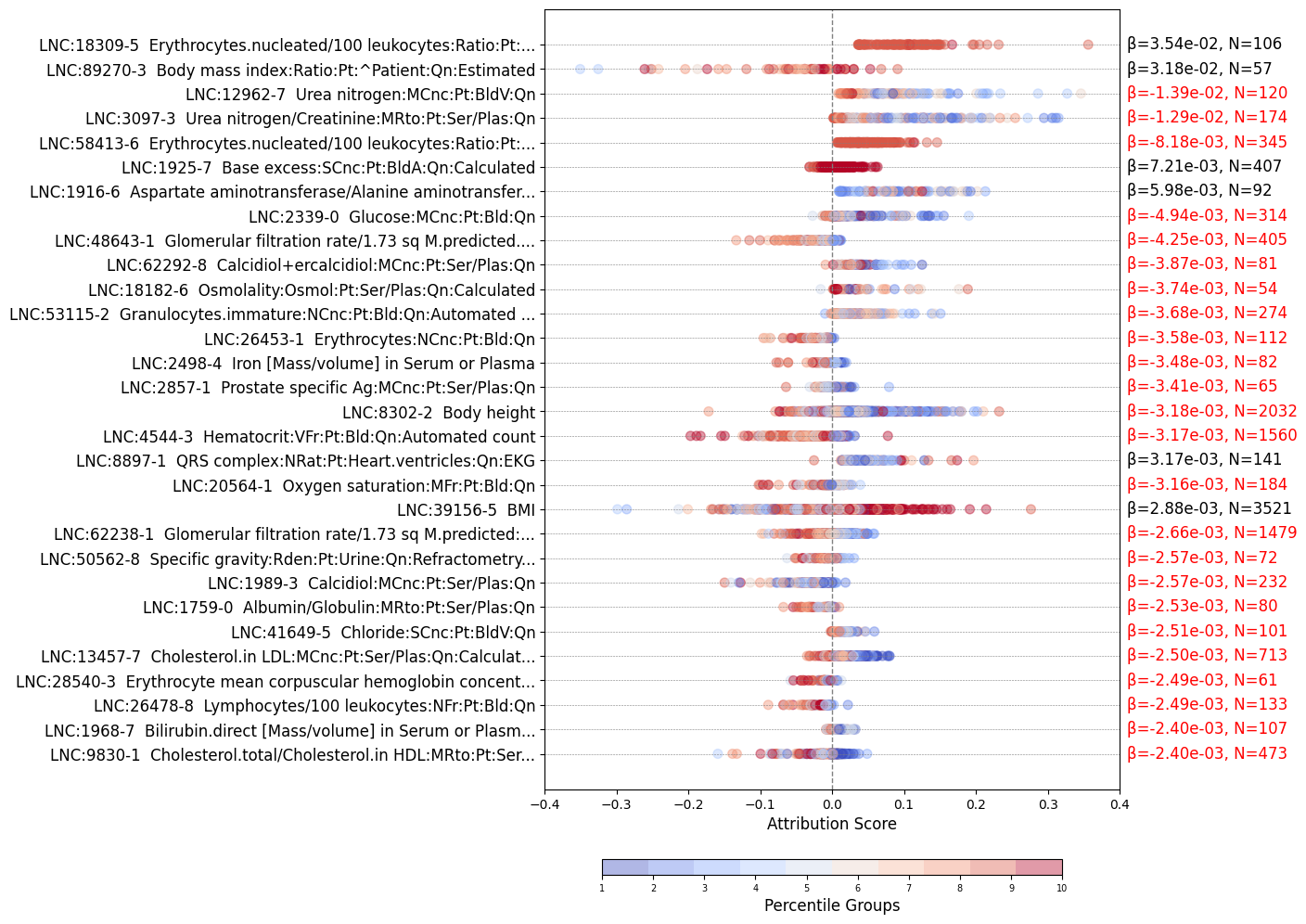}
 \caption{Top 30 laboratory test tokens with the highest regression coefficients for asthma loss of control.
 Each dot represents a lab test from a patient, with values indicated by color in decile groups.
Negative regression coefficients are labeled in red to distinguish them from positive regression coefficients. Tokens have minimum frequency $\geq 40$.
 }
 \label{fig:lab_scores}

\end{figure}

An important aspect of BERT-LER in this study is that laboratory features can be examined through the same attribution framework used for the rest of the EHR sequence. 
Analysis of laboratory features associated with asthma worsening (loss of control) reveals several modest but clinically interpretable correlations spanning hematologic, metabolic, renal, and systemic physiologic domains (Figure~\ref{fig:lab_scores}). 
\rev{An important caveat is confounding by indication, or informed-presence bias: the mere ordering of a laboratory test is not random and often encodes care setting and acuity, since some panels (e.g., urea/creatinine, osmolality) are predominantly ordered in the inpatient setting while others are typical of routine outpatient care \cite{agniel2018biases, phelan2017informed}. As a result, attributions attached to the presence or magnitude of a lab token may partly reflect where and why the test was ordered rather than a direct physiological link to asthma worsening. The percentile-value ablation (Table~\ref{tab:performance_metrics}) helps disentangle these effects by isolating value magnitude from lab occurrence, but the directional lab associations below should still be read as hypothesis-generating.}

Higher body mass index (BMI) is positively correlated with asthma worsening, consistent with evidence that obesity worsens asthma control via mechanical restriction, systemic inflammation, and altered immune responses \cite{obesity}. 
These patterns are also reflected in the overall top tokens (Figure~\ref{fig:overall_scores}) and in comparison models (Figures~\ref{fig:retain_top_features}, \ref{fig:xgboost_top_features} in the Appendix).
Hematologic parameters also emerge as relevant signals. Elevated nucleated erythrocyte ratios are positively correlated with asthma worsening. 
The presence of nucleated red blood cells in peripheral blood is generally indicative of bone marrow stress, hypoxemia, or systemic inflammation \cite{nrbc_respiratory_distress}. 
Chronic or intermittent hypoxic stress is a known feature of poorly controlled asthma and may explain this association. 
Additionally, base excess shows a positive correlation with loss of asthma control, potentially reflecting compensatory metabolic changes associated with chronic ventilatory strain or impaired gas exchange in severe obstructive airway disease.
Similar acid–base changes, including mixed respiratory and metabolic disturbances, have been documented in patients with severe asthma exacerbations and chronic respiratory compromise \cite{acid_base}.

In contrast, several laboratory markers demonstrate negative correlations with asthma worsening. 
Renal function markers, including blood urea nitrogen and estimated glomerular filtration rate (eGFR), are inversely associated with asthma loss of control. 
While not directly involved in airway pathophysiology, eGFR reflects kidney function and overall systemic health. 
Lower eGFR has been reported in individuals with asthma, potentially due to chronic inflammation, hypoxia, and shared cardiometabolic comorbidities~\cite{gfr}. 
In this context, preserved renal function may reflect lower comorbidity burden and systemic inflammation, both associated with better asthma control.

Markers related to oxygenation and red blood cell mass also exhibit negative correlations. 
Higher hematocrit, erythrocyte count, and oxygen saturation are associated with reduced likelihood of asthma worsening, consistent with clinical observations that low red blood cell count, chronic hypoxemia, and impaired oxygen delivery are features of uncontrolled asthma and severe exacerbations \cite{asthma_oxygen, asthma_hemoglobin}. 
Several metabolic and inflammatory biomarkers further support this pattern:
higher serum vitamin D, iron levels, and HDL-related lipid ratios show modest negative correlations with asthma worsening.
Vitamin D has been linked to immune regulation and airway inflammation, with deficiency associated with increased exacerbation risk \cite{vitamin_d}. Similarly, iron sufficiency and favorable lipid profiles are associated with reduced systemic inflammation, which may indirectly protect against asthma instability \cite{iron}.

\section{Discussion}

We situate BERT-LER as a contribution to predictive modeling and foundation-style EHR representation learning, with a set of benchmarked clinical prediction tasks from the EHRShot dataset and a second study on a real-world asthma severity progression protocol.
Relative to prior EHR transformers that emphasize diagnoses and procedures, BERT-LER combines percentile-based laboratory value discretization with Integrated Gradients attributions in a single EHR modeling pipeline.
In addition to improving performance over simpler ML approaches, BERT-LER reduces reliance on expert-curated feature engineering by processing the full EHR dataset as-is, enabling generalization to new prediction tasks without redesigning task-specific features. 
Such reliance also limits a model's ability to generalize to novel diseases where key factors are not yet well understood.

In terms of explainability, Integrated Gradients attributions \rev{often reflect known risk factors} for asthma loss of control, anemia-related lab results, and more, and are largely consistent with the literature and with the RETAIN/XGBoost comparison models, as detailed in the Results.
For anemia, the model recovers established age, sex, and comorbidity (gut disorder) associations.
For asthma loss of control, top-ranked predictors span disease-severity codes, controller/reliever medication use, comorbidities such as COPD, obesity, demographic disparities, and laboratory markers of systemic stress versus physiologic resilience, in each case aligning with established risk factors and comparison models.
\rev{This pattern is consistent with the lab-ablation comparison, where removing percentile-value embeddings while retaining lab tokens leads to lower asthma prediction performance, suggesting that both the occurrence and the magnitude of laboratory measurements contribute to BERT-LER's predictive gains.}

\paragraph{Limitations}
Despite these strengths, several challenges remain. Our approach does not explicitly model potential confounders, and this is especially relevant for laboratory tokens: \rev{because test ordering is driven by clinical concern and care setting, lab presence and value may partly encode acuity or care setting rather than direct disease physiology (confounding by indication and informed-presence bias) \cite{agniel2018biases, phelan2017informed}, so attributions should be read as predictive associations, not causal effects.}
\rev{Integrated Gradients itself can distribute importance across correlated tokens and does not make feature interactions explicit; more broadly, our validation is qualitative rather than a formal quantitative test of attribution faithfulness or stability, which, along with identity-versus-value analyses, we leave to future work.}
\rev{The shared percentile embedding also means lab-specific meaning is learned in later layers rather than at the input, which may limit resolution for tests whose interpretation is strongly non-monotonic in percentile.}
In addition, the model cannot interpret novel clinical codes absent from training, limiting generalization to rare or newly introduced concepts.
\rev{Finally, asthma performance is lower than on several EHRShot tasks, likely reflecting the difficulty of forecasting rare, only partially predictable progression events and the smaller asthma fine-tuning cohorts.}
Future work could integrate confounder-invariant feature learning, as explored in medical imaging \cite{Zhao2020}, to further improve robustness and interpretability.
Overall, BERT-LER shows that combining laboratory-aware representation with token-level explanations can support both predictive performance and clinical interpretability, offering a foundation for more generalizable, trustworthy clinical risk prediction.

\section{Code Availability}
We have made our training package available at \rev{\url{https://github.com/Sanofi-Public/CLM-LER}}.
We have made our general-purpose explainability software available at \rev{\url{https://github.com/Sanofi-Public/Clinical-BERT-Explainability}}.

\newpage
\bibliography{references}

\newpage
\appendix
\setcounter{figure}{0}
\setcounter{table}{0}
\renewcommand{\thefigure}{A.\arabic{figure}}
\renewcommand{\thetable}{A.\arabic{table}}
\section{Data Source and Governance}
\label{section:data_governance}

All EHR data used for pre-training and custom benchmarking were obtained from \rev{TriNetX Dataworks} \cite{trinetx}. \rev{TriNetX Dataworks} provides access to approximately 75 million patients, including diagnoses, procedures, medications, and laboratory values.

This retrospective study using \rev{TriNetX Dataworks} is exempt from informed consent. 
The data reviewed are a secondary analysis of existing data, do not involve intervention or interaction with human subjects, and are de-identified according to the de-identification standard defined in Section §164.514(a) of the HIPAA Privacy Rule. The de-identification process is attested through a formal determination by a qualified expert as defined in Section §164.514(b)(1) of the HIPAA Privacy Rule. This determination was refreshed in December 2020.

\section{Asthma Study Design}
\label{section:asthma_study_design}

\begin{figure}[h]
    \centering
    \includegraphics[width=0.8\linewidth]{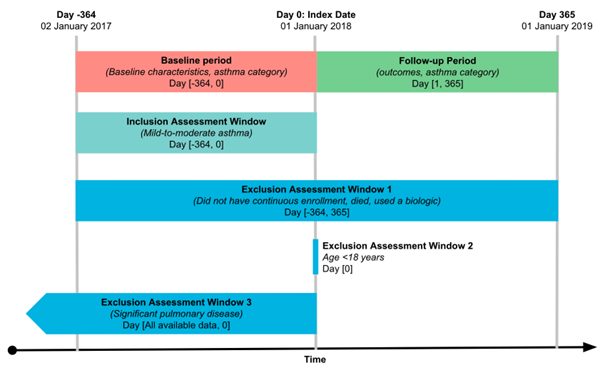}
\caption{The study index date of January 1, 2018 was selected to ensure that all patients had one year of baseline data and one year of follow-up data prior to the COVID-19 pandemic, which impacted asthma care and management \cite{Philip2022, Lin2022}. The figure illustrates the index date along with the baseline and follow-up observation windows used for cohort construction.}
    \label{fig:study_design}
\end{figure}

This study was designed to develop machine learning models predicting progression from mild-to-moderate asthma to more severe disease states using real-world electronic health record (EHR) data. Disease progression was operationalized using three mutually exclusive asthma profiles (Exacerbation, Symptoms without Exacerbation, Controlled), defined during both a baseline and follow-up period (see Section~\ref{section:asthma_profile_def}). Figure~\ref{fig:study_design} illustrates the study index date and observation windows.

\subsection{Study Timeline}
The following time-periods were used to identify the study population:
\begin{itemize}
    \item \textbf{Study Period:} January 2, 2017 – January 1, 2019
    \item \textbf{Patient Identification Period:} Patients with mild-to-moderate asthma were identified from January 2, 2017 (Day -364) through January 1, 2018 (Day 0)
    \item \textbf{Index Date:} January 1, 2018
    \item \textbf{Baseline Period:} The 365 days prior to and including the index date (January 2, 2017 – January 1, 2018)
    \item \textbf{Follow-up Period:} The 365 days after the index date (January 2, 2018 – January 1, 2019), allowing for a fixed one-year follow-up
\end{itemize}

\subsection{Inclusion and Exclusion Criteria}
\label{section:incl_excl_criteria}
\textbf{Inclusion Criteria}

Patients were included if they had mild-to-moderate asthma, defined as:
\begin{itemize}
    \item At least one inpatient or outpatient diagnosis of asthma in the primary position during the 365-day baseline period
    \item Use of any therapy for GINA Step 2 through Step 4 asthma according to the 2017 guidelines \cite{gina2017} during the 365-day baseline period
    \begin{itemize}
        \item \textbf{Step 2 therapies:} low-dose ICS, leukotriene receptor antagonist (LTRA), low-dose theophylline
        \item \textbf{Step 3 therapies:} low-dose ICS/LABA, medium-dose ICS, high-dose ICS, low-dose ICS + LTRA, low-dose ICS + theophylline
        \item \textbf{Step 4 therapies:} medium-dose ICS/LABA, high-dose ICS/LABA, medium-dose ICS/LABA + tiotropium, high-dose ICS/LABA + tiotropium, high-dose ICS + LTRA, high-dose ICS + theophylline
    \end{itemize}
\end{itemize}

\textbf{Exclusion Criteria}

Patients were excluded if they met any of the following criteria:
\begin{itemize}
    \item Lack of continuous insurance enrollment during the 365-day baseline and follow-up periods ($\leq30$-day gaps allowed)
    \item Death during the 365-day baseline or follow-up period
    \item Age <18 years or missing age on the index date
    \item Evidence of significant pulmonary disease other than asthma prior to the index date, including: chronic obstructive pulmonary disease, interstitial lung disease or pulmonary fibrosis, sarcoidosis, pulmonary hypertension, bronchiectasis, and Churg-Strauss syndrome
    \item Use of a biologic therapy indicated for severe asthma (omalizumab, mepolizumab, benralizumab, reslizumab, dupilumab) at any point during the baseline or follow-up period
\end{itemize}

\rev{Two exclusion criteria---death and initiation of a severe-asthma biologic---are defined using follow-up information. Because both events may be related to baseline severity and to the predicted outcome, conditioning on their absence is a form of selection on a post-baseline variable that can bias the cohort toward less severe trajectories and attenuate associations with the most severe outcomes. These exclusions were retained to keep the target cohort aligned with the mild-to-moderate, biologic-na\"ive population of clinical interest and to avoid an ambiguous label period for patients who die during follow-up. Treating these events as censoring or competing risks within a time-to-event formulation is a natural direction for future work.}

\subsection{Asthma Profile Definitions}
\label{section:asthma_profile_def}

Patients meeting the inclusion/exclusion criteria were classified into one of three mutually exclusive asthma profiles during two separate periods: the 365-day baseline period and the 365-day follow-up period.

\begin{itemize}
    \item \textbf{Profile 1: Exacerbation}
    \begin{itemize}
        \item Inpatient visit with asthma as the primary, admitting, or discharge diagnosis
        \item Emergency department visit with asthma as the primary diagnosis
        \item Oral corticosteroid prescription ($\geq3$ days but <30 days) within 7 days of an asthma diagnosis
        \item Outpatient procedure indicating systemic corticosteroid administration (oral or injectable) within 7 days of an asthma diagnosis
    \end{itemize}

    \item \textbf{Profile 2: Symptoms without Exacerbation}
    \begin{itemize}
        \item Step-up in GINA therapy (e.g., Step 2 → 3, Step 3 → 4)
        \item Receipt of $\geq6$ SABA rescue inhaler prescriptions
        \item Diagnosis of uncontrolled asthma symptoms (dyspnea, wheezing, chest pain, or cough) at an outpatient visit with an asthma diagnosis in any position
        \item Oral corticosteroid prescription with <3-day supply
        \item $\geq2$ outpatient visits with an asthma diagnosis code in any position
    \end{itemize}

    \item \textbf{Profile 3: Controlled}
    \begin{itemize}
        \item No events qualifying for Profile 1 or 2
    \end{itemize}
\end{itemize}

\noindent Patients experiencing both Profile 1 and Profile 2 events during the same period were classified as Exacerbation (Profile 1).
\section{Model Pretraining}
\label{supp:model_pretraining}
Each model was pretrained using MLM, scanning over a set of hyperparameters.

\begin{table}[h]
\caption{Pretraining hyperparameter sweep, with the selected parameters highlighted. The same parameters were used for both Med-BERT and BERT-LER. Pretraining was performed on a single A10 GPU with 24 GB of VRAM. The best-performing model saw 20 million unique patient records during training, slightly more than one epoch of data.}
\centering
\begin{small}
\begin{tabular}{ll}
\toprule
\textbf{Name} & \textbf{Values}  \\
\midrule
Learning Rate & \textbf{5e-5}, 1e-5, 5e-6  \\
Effective Batch Size & 16, \textbf{32} \\
\bottomrule
\end{tabular}
\end{small}
\end{table}

\begin{table}[h]
\caption{Other training parameters for the model. Early stopping monitored the validation loss, stopping training after no improvement over 3 validation steps and at least one completed epoch. During hyperparameter tuning, models stopped between 1 and 2 epochs, having seen 20 to 40 million unique patient medical histories.}
\centering
\begin{small}
\begin{tabular}{ll}
\toprule
\textbf{Name} & \textbf{Values} \\
\midrule
number of layers & 12 \\
number of attention heads & 12 \\
attention dropout & 0.15 \\
positional encoding & relative key query \\
embedding dimension in embedding layer & 768 \\
embedding dimension in hidden layers & 768 \\
fp16 & True \\
N epochs & 5 \\
max positional embedding & 1825 \\
training steps per validation step & 10000 \\
early stopping patience & 3 validation steps \\
early stopping delay & 1 epoch \\
Fraction of tokens masked & 0.15 \\
Fraction of masked tokens replaced by a random token & 0.1 \\
Fraction of masked tokens replaced by the original token & 0.1 \\
Fraction of masked tokens replaced by mask token & 0.8 \\
Validation set fraction & 1.5 \% \\
\bottomrule
\end{tabular}
\end{small}
\end{table}

\section{Fine-tuning/Classification Head Details}
\label{supp:finetune_head}

\rev{For fine-tuning, every input sequence begins with a special \texttt{[CLS]} token whose final-layer embedding summarizes the sequence. For an $M$-class task, classification is performed by a single fully connected layer that maps this \texttt{[CLS]} embedding to $M$ logits, trained jointly with the pretrained encoder.}

\section{Confidence Interval Estimation}
\label{supp:bootstrap}

\rev{All reported confidence intervals are obtained by bootstrapping the held-out test set with 1{,}000 resamples drawn with replacement; we report the metric on the full test set together with the 2.5th and 97.5th percentiles of the bootstrap distribution as the 95\% confidence interval.}

\newpage

\section{Laboratory Value Representation and Masking}
\label{supp:lab_representation}

\rev{BERT-LER represents each laboratory measurement at a single sequence position using two embeddings that are summed: a lab-code embedding that encodes the test identity and a shared percentile embedding that encodes the value rank (Section~\ref{section:bert_train}). An alternative is to assign a distinct token to every laboratory-test-by-percentile combination. With $N$ laboratory tests and $M$ percentile bins, the factorized representation requires approximately $N + M$ embeddings, whereas the per-combination representation requires $N \times M$ embeddings, substantially increasing parameter count and memory use. This gap widens under finer value resolution: increasing the number of bins from $M$ to $M'$ adds only $M' - M$ embeddings to the factorized table, but adds $N \times (M' - M)$ embeddings to the per-combination table. Because the percentile embedding is shared across tests, the clinical meaning of a given percentile (e.g., a ``high'' value) differs by test, and the model learns these test-specific interactions in later layers; this trade-off is noted as a limitation in the Discussion. Joint continuous--categorical encodings are a further alternative to binning \cite{binning_joint_embeddings}, and the lab-value ablation in the main text isolates the contribution of the percentile-value component within the present design.}

\rev{The lowest and highest bins are open-ended tail bins (effectively \((-\infty,\mathrm{P}10]\) and \([\mathrm{P}90,+\infty)\)), so all low-end and high-end values are captured without separate outlier handling.}
\rev{The percentile cut points are fixed population-level reference statistics estimated once over the full reference population, independent of any task label or train/validation/test partition, and are reused unchanged across all tasks.}
\rev{Missing values require no special handling: because the input is an event sequence, missing data are handled implicitly, in that an event that was never recorded simply does not appear as a token, and no imputation is applied for absent diagnoses, medications, or laboratory results.}
Following this procedure, each patient’s data is represented as three input arrays: 

\rev{During masked language modeling pre-training, lab-code tokens and their percentile IDs are not masked jointly. When a laboratory-code token is masked, the model still observes the associated percentile-bin embedding and predicts the masked laboratory code from the surrounding clinical context together with the retained percentile information.}

\newpage

\section{EHRShot/CLMBR Vocabulary}
\label{supp:clmbr_vocab}

\rev{In both EHRShot (canonical benchmark splits with defined prediction times) and our asthma tasks (Appendix, Study Timeline), only events up to and including the index date are fed to the model, while labels are derived solely from the post-index follow-up window. The laboratory tokens used as inputs are therefore strictly pre-index and cannot contain the outcome being predicted, so the strong lab-task performance is not an artifact of target leakage.}

\rev{For comparison with the BERT-LER and MED-BERT vocabularies in Table~\ref{tab:vocab_dist}, Table~\ref{tab:clmbr_vocab} reports the composition of the released CLMBR-T-base vocabulary, reproduced from its model card \cite{ehrshot}. CLMBR is trained with an autoregressive next-code objective, and its vocabulary is the top $65{,}536$ codes (the maximum value of a \texttt{uint16}) drawn from $21$ OMOP source ontologies, ranked by global frequency in the Stanford pre-training data; codes outside this set are dropped. Two differences are relevant to our experiments. First, CLMBR encodes diagnoses through SNOMED concepts, whereas our pre-training and asthma cohorts use ICD10CM and ICD9CM, which is one reason the released CLMBR model is not directly portable to the asthma cohort without substantial concept remapping. Second, laboratory tests dominate the CLMBR vocabulary (LOINC, $37{,}590$ codes), reflecting the breadth of laboratory coverage in the benchmark. Note that CLMBR's LOINC code vocabulary contains a unique token for each lab-test and lab-percentile combination. This creates $\sim$10x more lab test tokens than the BERT-LER model. We chose to encode these percentile values as a separate input.}

\begin{table}[h]
\caption{\rev{Composition of the released CLMBR-T-base vocabulary by OMOP source ontology, reproduced from the model card \cite{ehrshot}. Purpose labels are approximate and parallel Table~\ref{tab:vocab_dist}; SNOMED in OMOP can span conditions, procedures, and observations.}}
\label{tab:clmbr_vocab}
\centering
\begin{small}
\rev{
\begin{tabular}{l|l|r}
\toprule
Source ontology & Purpose & \# codes \\
\midrule
LOINC & Lx & 37{,}590 \\
SNOMED & Dx / clinical & 18{,}174 \\
RxNorm & Rx & 4{,}678 \\
CPT4 & Pr & 3{,}730 \\
CARE\_SITE & Other & 396 \\
RxNorm Extension & Rx & 255 \\
ICD10PCS & Pr & 233 \\
ICD9Proc & Pr & 196 \\
Cancer Modifier & Other & 88 \\
HCPCS & Pr & 54 \\
ICDO3 & Dx / oncology & 52 \\
CVX & Vx & 41 \\
Domain & Other & 27 \\
Visit & Other & 6 \\
Race & Demo & 5 \\
OMOP Extension & Other & 3 \\
Gender & Demo & 2 \\
Ethnicity & Demo & 2 \\
CMS Place of Service & Other & 2 \\
Medicare Specialty & Other & 1 \\
Condition Type & Other & 1 \\
\midrule
Total & & 65{,}536 \\
\bottomrule
\end{tabular}
}
\end{small}
\end{table}

\newpage

\section{Comparison Methods for Asthma Tasks}
\label{supp:count_comparison}

To evaluate BERT-LER on asthma tasks and benchmark it against prior methods, we trained a diagnosis-only BERT model (Section~\ref{section:bert_train}) and several supervised machine learning models on the three predefined tasks. Specifically, we compared BERT-LER to:

\begin{itemize}
    \item \textbf{Med-BERT (Diagnosis-only BERT model):} A large-scale pre-trained EHR language model that treats medical records as sequences of clinical tokens but uses only diagnoses data \cite{Rasmy2021medbert}.
    \item \textbf{RETAIN:} A reverse-time attention neural network for interpretable prediction on longitudinal EHR sequences. RETAIN applies visit-level and variable-level attention to highlight key clinical events while maintaining competitive predictive performance \cite{retain}.
    \item \textbf{XGBoost with Count-Based Featurization:} A tree-based gradient boosting algorithm known for high performance on structured datasets \cite{xgboost}.
    \item \textbf{Logistic Regression with Count-Based Featurization:} A standard linear baseline widely used in clinical classification tasks \cite{scikit-learn}.
\end{itemize}

For XGBoost and Logistic Regression, we employed count-based featurization. In this approach, each patient's medical record was transformed into a vector encoding the number of occurrences of each distinct medical event observed within the specified timeline. Aggregating these vectors across all patients yielded a high-dimensional, sparse feature matrix.
\rev{Concretely, the time series of events for each patient is collapsed into a fixed-length vector indexed by the same clinical-event vocabulary used by the language models, where each entry counts how many times that event occurs in the patient's history. For example, with a vocabulary of [Asthma, Diabetes], a patient with one asthma diagnosis and three diabetes diagnoses is represented as [1, 3]. Importantly, the count-based baselines have access to laboratory values, not only to the fact that a test was ordered: each laboratory event is keyed by both its test code and its percentile bin (e.g., a creatinine result in the 70--80th percentile and one in the 90--100th percentile are distinct features), so the value information available to BERT-LER is also available to these baselines. This differs only in representation from BERT-LER, which keeps a single token per laboratory code and encodes the percentile through a separate embedding (hence the smaller laboratory vocabulary in Table~\ref{tab:vocab_dist}). RETAIN uses the same percentile-keyed laboratory vocabulary but preserves the longitudinal sequence structure rather than collapsing it into counts.} To address the resulting dimensionality, we optionally applied Truncated Singular Value Decomposition (TSVD) \cite{scikit-learn}, a dimensionality reduction method analogous to Principal Component Analysis (PCA) but optimized for sparse data, to reduce the feature space while retaining the majority of the variance. We performed hyperparameter search using scikit-learn’s \texttt{ParameterGrid} function \cite{scikit-learn}, with the selected configurations reported in Table~\ref{tab:asthma_model_hyperparam}. After model selection, we applied bootstrapping to obtain the final performance estimates.

\begin{table}[h]
\caption{Hyperparameters used for the models trained on the asthma prediction tasks.}
\begin{footnotesize}
\setlength{\tabcolsep}{4pt}
\begin{tabular}{l|l|l|l|l|l}
\toprule
\multirow{2}{*}{\textbf{Model}} & \multirow{2}{*}{\textbf{Name}} & \multirow{2}{*}{\textbf{Values}} & \multicolumn{3}{c}{\textbf{Task \& Best Value}} \\
 &  &  & Loss-of-control & Multiclass & Exacerbation \\
\midrule
\multirow{3}{*}{BERT-LER} & Learning Rate & 5e-5, 1e-5, 5e-6 & 1e-5 & 5e-6 & 5e-6 \\
 & Gradient Accumulation Steps & 1, 2 & 1 & 1 & 1 \\
 & Attention Dropout Probability & 0.0, 0.2, 0.4 & 0.4 & 0.2 & 0.0 \\
\midrule
\multirow{3}{*}{Med-BERT} & Learning Rate & 5e-5, 1e-5, 5e-6 & 5e-6 & 1e-5 & 5e-6 \\
 & Gradient Accumulation Steps & 1, 2 & 1 & 1 & 1 \\
 & Attention Dropout Probability & 0.0, 0.2, 0.4 & 0.2 & 0.4 & 0.4 \\
\midrule
\multirow{3}{*}{RETAIN} & Learning Rate & 1e-5, 1e-4, 1e-3 & 1e-4 & 1e-3 & 1e-3 \\
 & Embedding Size & 128, 64 & 64 & 64 & 64 \\
 & Dropout & 0.0, 0.2, 0.4 & 0.4 & 0.2 & 0.0 \\
\midrule
\multirow{4}{*}{XGBoost} & TruncatedSVD \# Components & None, 50, 100, 150 & None & None & None \\
 & Learning Rate & 0.05, 0.1 & 0.1 & 0.1 & 0.05 \\
 & Max Depth & 3, 5, 7 & 7 & 5 & 3 \\
 & Gamma & 0.0, 0.5, 1.0 & 0.5 & 0.5 & 0.0 \\
\midrule
\multirow{2}{*}{Logistic Regression} & TruncatedSVD \# Components & None, 50, 100, 150 & 150 & 100 & 150 \\
 & C & 0.001, 0.01, 0.1, 1, 10 & 1 & 0.001 & 0.01 \\
\bottomrule
\end{tabular}
\end{footnotesize}
\label{tab:asthma_model_hyperparam}
\end{table}

\newpage

\section{Explainability Attribution Score Aggregation} 
\label{section:bert_attribution_aggr}
Attribution scores were calculated for each token for every patient in the test set. 
These individual scores were then averaged across the entire set and ranked by magnitude to 
identify the tokens with the highest overall attribution scores in the population.
To ensure that the aggregated attribution scores reflect meaningful insights, 
an adjustment was applied to account for infrequent tokens using the following formula:

\[
\text{adj.score} = \text{avg.score} \times \max\left(0, \left(1 - \frac{k}{n}\right)\right)
\]

where \(k\) is a user-defined minimum frequency threshold and \(n\) represents the count of occurrences of the token. 
Tokens with a frequency less than or equal to \(k\) were assigned a score of 0, with rarer events receiving heavier penalties.

The reference baseline embedding for all non-special tokens, except demographic tokens (gender, age, region, race, ethnicity) and laboratory results tokens, 
is set to zero embeddings. For demographic tokens, since they always appear in fixed positions at the start of each input, 
the BERT model may pick up on this pattern and overemphasize their importance, leading to biased, elevated attribution scores. 
In this context, a zero embedding (which represents the absence of a token) is not a neutral baseline. 
Therefore, for each demographic category, we use the average embedding of tokens within that category as the baseline. 
For example, if "male", "female", and "unknown" are the tokens in the gender demographic category, 
the baseline reference embedding is the average of those three embeddings.
For laboratory results tokens, the baseline percentile embedding is set to the median percentile range.

Laboratory results tokens are treated differently due to their values being represented as ordinal percentile ranges.
We consider a lab test predictive of a task outcome when the test's percentile aligns with the attribution scores and spans a wide range of scores.
To measure this, we fit a linear regression with ordinal lab percentile bins (ten groups) as the independent variable and token attribution scores as the dependent variable, ranking tokens by the absolute regression coefficient.
Additionally, to compare their attribution scores against other types of tokens, we split each laboratory test token into three distinct tokens:
"token-LOW" representing values that fall in the 0-30\% percentile, 
"token-MODERATE" representing values that fall in the 30-70\% percentile, and
"token-HIGH" representing those who fall in the 70-100\% percentile.
Each of these token-level pairs is then averaged separately and adjusted for comparison with non-laboratory test tokens.

\newpage

\section{RETAIN Explanations}
\label{section:retain_top_features}

\begin{figure}[h]
    \centering
    \includegraphics[width=1\linewidth]{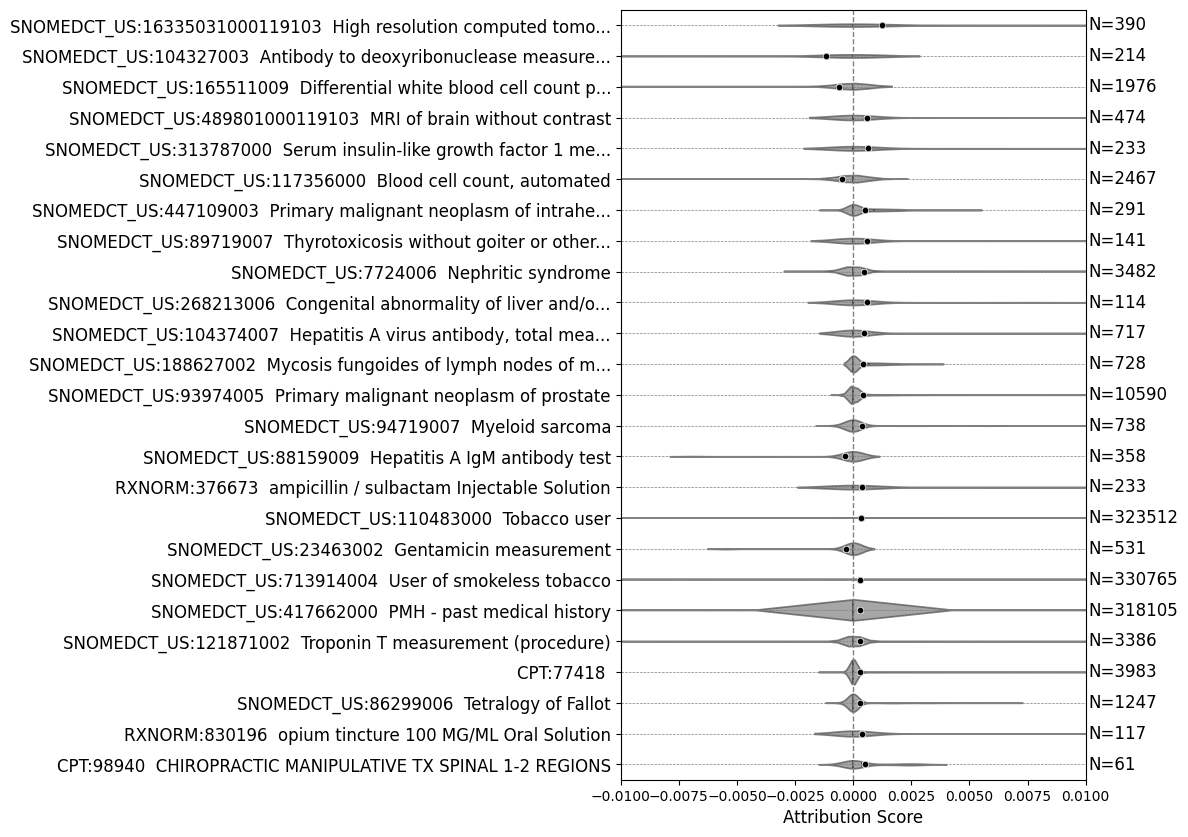}
    \caption{Top contributing factors from the RETAIN model for lab anemia prediction.}
    \label{fig:retain_lab_anemia}
\end{figure}

\newpage

\begin{figure}[h]
  \centering
  \includegraphics[width=1\linewidth]{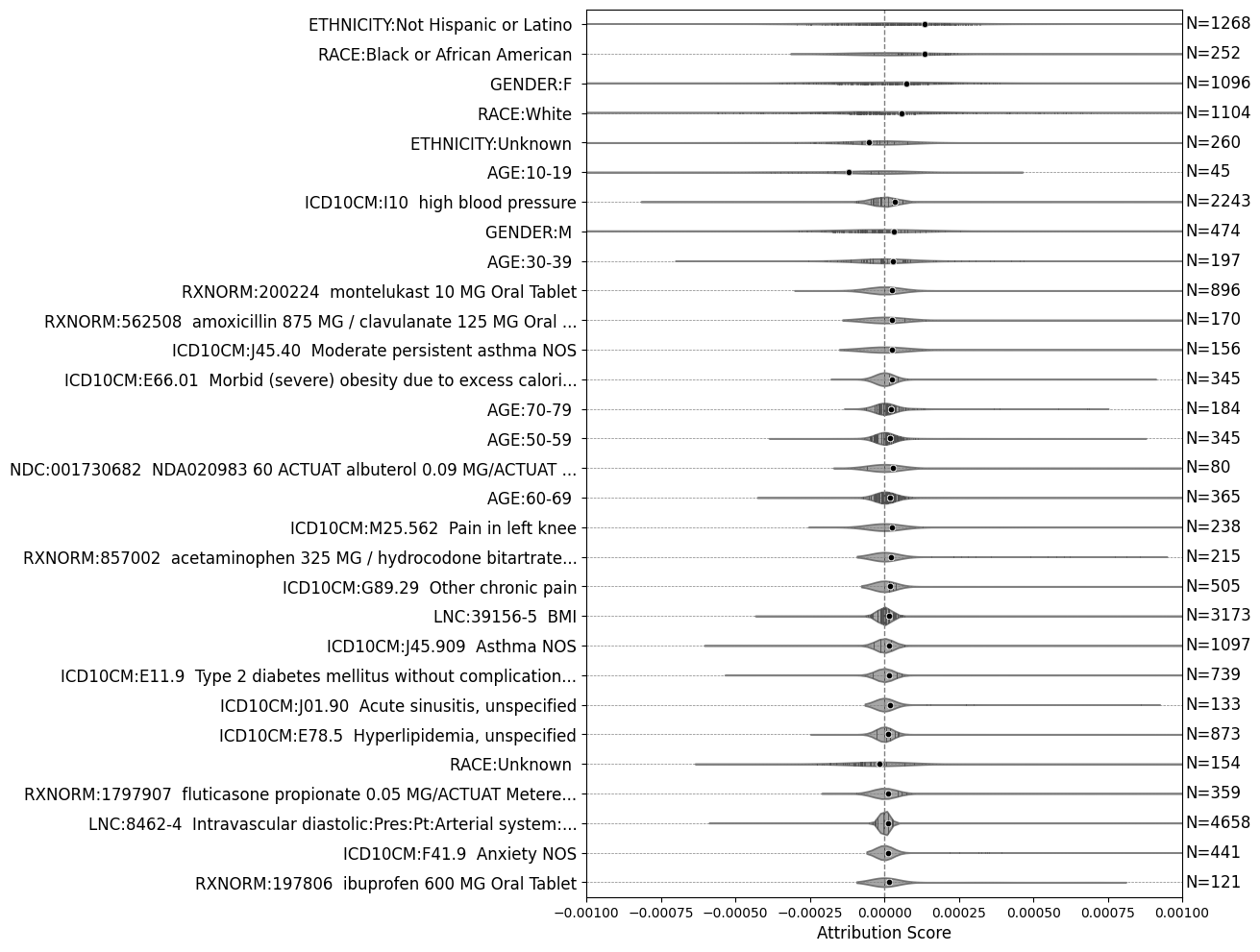}
    \caption{Top contributing factors identified by the RETAIN model for predicting asthma loss of control.}
  \label{fig:retain_top_features}
\end{figure}

\newpage

\section{XGBoost Explanations}
\label{section:xgboost_top_features}
\begin{figure}[h]
  \centering
  \includegraphics[width=1\linewidth]{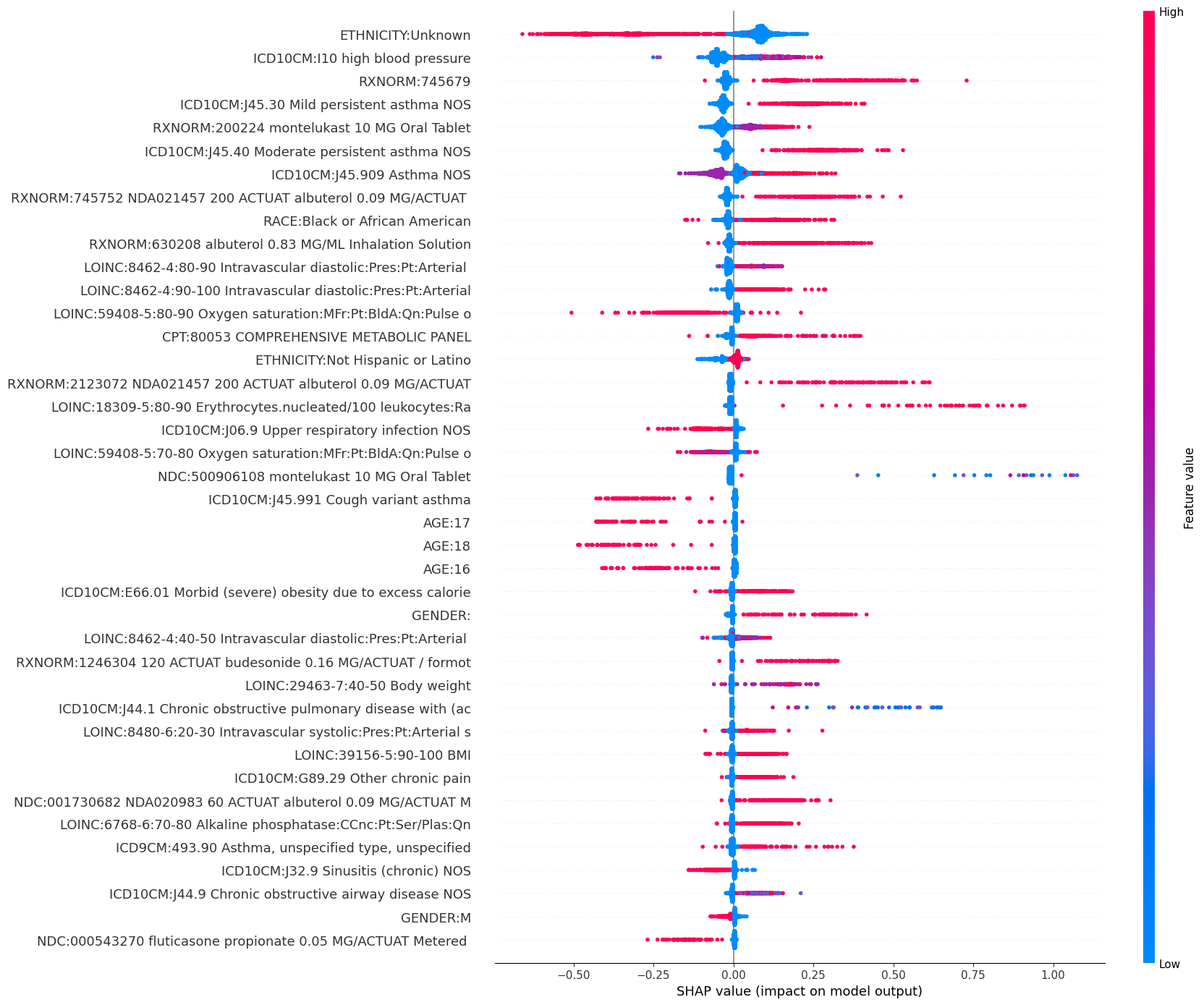}
    \caption{Top contributing factors identified by the XGBoost model for predicting asthma loss of control.}
  \label{fig:xgboost_top_features}
\end{figure}

\end{document}